\pdfoutput=1

\documentclass[11pt]{article}

\usepackage[]{acl}

\usepackage{times}
\usepackage{latexsym}

\usepackage[T1]{fontenc}

\usepackage[utf8]{inputenc}

\usepackage{microtype}
\usepackage{bm}
\usepackage{graphicx}
\usepackage{amsmath}
\usepackage{multirow}
\usepackage{cleveref}
  \crefname{Figure}{Fig.}{Fig.}
\usepackage{xspace}
\usepackage{xcolor}
\usepackage{enumitem}
\usepackage{placeins}
\usepackage{booktabs}
\usepackage{soul}

\newcommand{\ours}{{QC4QA}\xspace}

\title{Domain Adaptation for Question Answering via Question Classification}

\author{Zhenrui Yue \and Huimin Zeng \and Ziyi Kou \and Lanyu Shang \and Dong Wang \\
  School of Information Sciences \\
  University of Illinois Urbana-Champaign \\
  \texttt{\{zhenrui3, huiminz3, ziyikou2, lshang3, dwang24\}@illinois.edu}}

\begin{document}
\maketitle
\begin{abstract}
Question answering (QA) has demonstrated impressive progress in answering questions from customized domains. Nevertheless, domain adaptation remains one of the most elusive challenges for QA systems, especially when QA systems are trained in a source domain but deployed in a different target domain. In this work, we investigate the potential benefits of question classification for QA domain adaptation. We propose a novel framework: Question Classification for Question Answering (\ours). Specifically, a question classifier is adopted to assign question classes to both the source and target data. Then, we perform joint training in a self-supervised fashion via pseudo-labeling. For optimization, inter-domain discrepancy between the source and target domain is reduced via maximum mean discrepancy (MMD) distance. We additionally minimize intra-class discrepancy among QA samples of the same question class for fine-grained adaptation performance. To the best of our knowledge, this is the first work in QA domain adaptation to leverage question classification with self-supervised adaptation. We demonstrate the effectiveness of the proposed \ours with consistent improvements against the state-of-the-art baselines on multiple datasets.
\end{abstract}

\section{Introduction}
\label{sec:introduction}

Question Answering~(QA) or Reading Comprehension~(RC) refers to the task of extracting answers from given context paragraphs based on input questions. QA systems predict the start and end positions of possible answer spans in given context documents upon input questions. In recent studies, QA systems have achieved significant improvements with transformer models and large-scale datasets~\cite{rajpurkar-etal-2016-squad, devlin-etal-2019-bert, yue-etal-2022-c}.

\begin{figure}[t]
  \centering
  \includegraphics[width=1.0\linewidth]{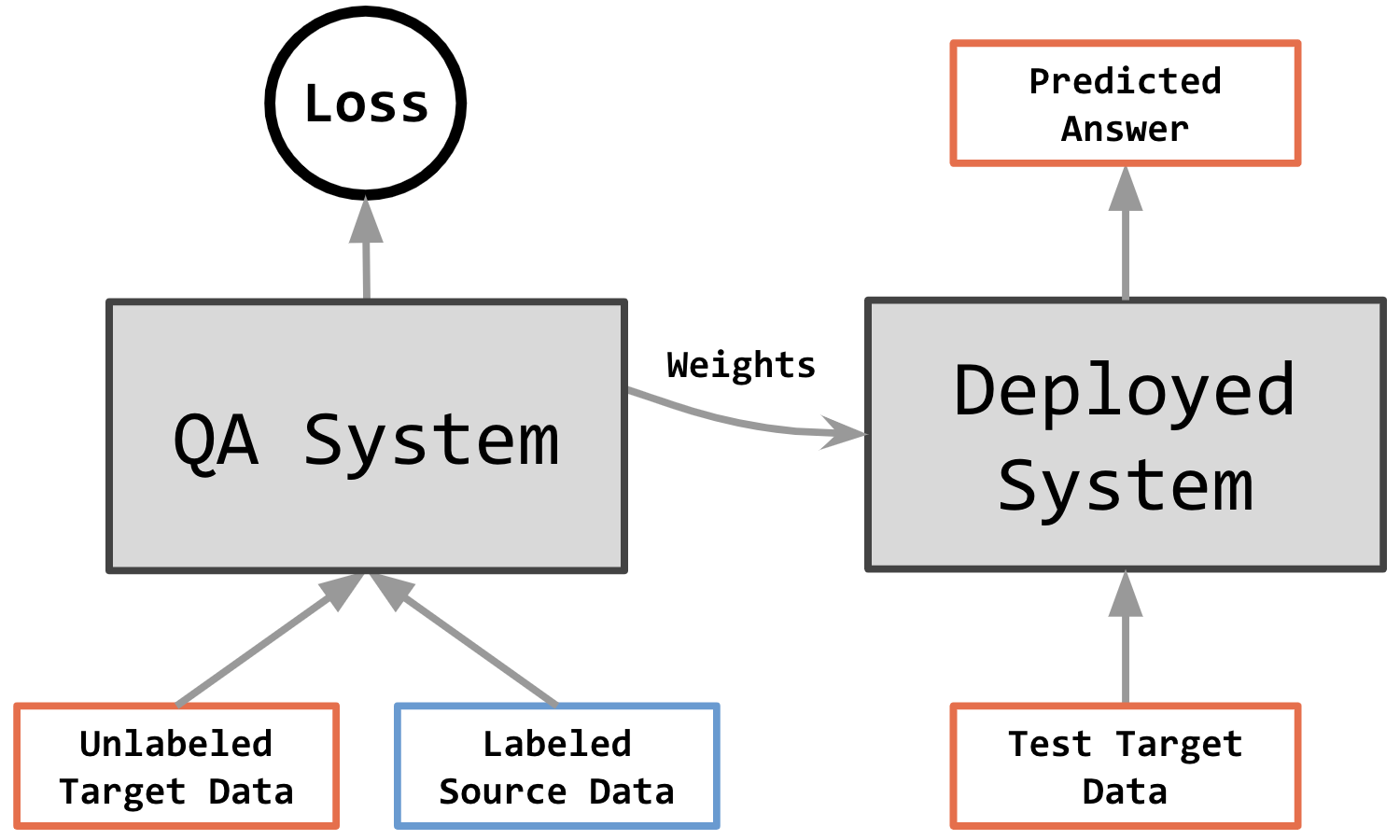}
  \caption{Overview for QA domain adaptation. A QA model is trained with labeled source data and unlabeled target data. The resulting QA system is deployed to answer target questions.}
  \label{fig:setting}
\end{figure}

Once deployed, QA systems often experience performance deterioration upon user-generated questions. Such performance drops can be traced back to domain shifts in two input elements: (1)~User-generated questions are syntactically more diverse and thus, different from the training QA pairs; (2)~The context domain of test-time input (target domain) can oftentimes diverge from the training corpora (source domain), e.g., from news snippets to biomedical articles~\cite{hazen2019towards, fisch-etal-2019-mrqa, miller2020effect}. 

To alleviate the performance issue in QA domain adaptation, several approaches have been proposed to reduce the discrepancy between the source and target domains. Integrating labeled target QA pairs in training can effectively improve the QA system in answering out-of-domain questions~\cite{kamath-etal-2020-selective, shakeri-etal-2020-end, yue-etal-2021-contrastive, yue-etal-2022-synthetic}, where the target data can be human-annotated QA pairs or synthetic data using question generation methods. When only unlabeled questions are available (see \Cref{fig:setting}), another possible approach is to reduce inter-domain discrepancy via domain-adversarial training~\cite{lee-etal-2019-domain}. Combined with pseudo labeling, QA systems demonstrate improved generalization in answering target domain questions~\cite{cao2020unsupervised}.

Nevertheless, previous methods either require large amounts of annotated target data or extensive computing power~\cite{lee-etal-2019-domain, cao2020unsupervised, yue-etal-2021-contrastive, yue-etal-2022-synthetic}. Additionally, different types of QA pairs and their distributional changes are not taken into account. As a result, existing approaches are less effective for adapting QA systems to an unseen target domain. In this paper, we propose a domain adaptation framework for QA: \emph{question classification for question answering} (\ours). Unlike existing methods, we innovatively adopt a question classification (QC) model to classify input questions from both the source and target domains into different question classes. Moreover, we pseudo label the target data using a pretrained QA system and perform distribution-aware sampling to build mini-batches that resemble the target question distribution. In the adaptation stage, we propose a self-supervised adaptation framework to minimize the domain gap, in which inter-domain and intra-class discrepancies are simultaneously regularized. This is in contrast to existing baselines (e.g., domain adversarial adaptation methods) where the source data is solely used for training but without explicitly accounting for domain shifts and question distribution changes~\cite{lee-etal-2019-domain, cao2020unsupervised, yue-etal-2021-contrastive}. To the best of our knowledge, \ours is the first work that combines question classification and self-supervised adaptation for learning domain-invariant representation in QA domain adaptation.

\emph{Our main contributions} are as follows\footnote{Our implementation is publicly available at https://github.com/Yueeeeeeee/Self-Supervised-QA.}:
\begin{enumerate}
\item We propose \ours for QA domain adaptation. \ours innovatively adopts question classification to identify question types (classes) of the source and target QA pairs for intra-class discrepancy reduction.
\item Our \ours can be combined with supervised QC or unsupervised clustering. In the latter case, we show that \ours can transfer knowledge to the target domain even without additional model or annotation.
\item We design a distribution-aware sampling strategy and an objective function that incorporates MMD distances for minimizing inter-domain and intra-class discrepancies to transfer knowledge to the target domain.
\item We demonstrate the effectiveness of \ours, where \ours consistently outperforms state-of-the-art baselines by a significant margin.
\end{enumerate}

\section{Related Work}
\label{sec:related_work}
QA systems have achieved significant improvements in extracting answers upon input context and questions. However, trained QA systems are known to experience performance drops when context paragraphs and questions diverge from the training corpora. That is, when domain shifts exist between the training and test distributions~\cite{hazen2019towards, fisch-etal-2019-mrqa, miller2020effect, zeng2022attacking}.

To adapt QA systems for domain changes, methods for QA domain adaptation have been proposed in two different settings: (1)~Access to contexts and QA pairs from the target domain. Here, partial access to target data is provided, or a question generation model is introduced for producing synthetic QA pairs. The target data is then used to train and improve adaptation performance~\cite{shakeri-etal-2020-end, yue-etal-2021-contrastive}; (2)~Access to context paragraphs and unlabeled input questions from the target domain. Here, unsupervised or self-supervised adaptation can be used to improve the performance in the target domain~\cite{cao2020unsupervised}. In this paper, we focus on the latter setting and study QA domain adaptation with access to target contexts and unlabeled questions.

\textbf{Domain adaptation in computer vision}: Domain adaptation methods have been primarily studied for image classification problems. Such approaches focus on minimizing the representation discrepancy between the source and target distributions. Some methods design objective functions that encourage domain-invariant features in training~\cite{long2015learning, kang2019contrastive}. Other methods leverage domain-adversarial training with a discriminator to implicitly impose regularization when source and target features are distinguishable~\cite{tzeng2017adversarial, zhang2019domain}, with successful applications in various vision tasks~\cite{zhang2019transland, zhang2020hybrid, zhang2021migratable}.

\textbf{Domain adaptation in QA}: Various approaches are designed to improve QA performance by generating and refining synthetic QA pairs. Based on target contexts, question generation models are introduced to produce a surrogate dataset, which is used to train QA systems~\cite{kamath-etal-2020-selective, shakeri-etal-2020-end, yue-etal-2022-synthetic}. Contrastive adaptaion minimizes inter-domain discrepancy with question generation and maximum mean discrepancy (MMD) distances~\cite{yue-etal-2021-contrastive, yue2022contrastive}. When unlabeled questions are accessible, domain-adversarial training can be applied to reduce feature discrepancy between domains~\cite{lee-etal-2019-domain}. Pseudo labeling and iterative refinements of such labels can be used for improved joint training~\cite{cao2020unsupervised}. 

\textbf{Question classification (QC)}: Classifying questions of different types is a common task in natural language processing. One of the widely-used question taxonomy TREC divides questions into 6 coarse classes and 50 fine classes~\cite{li-roth-2002-learning}. Early machine learning methods perform QC with hand-crafted features~\cite{li-roth-2002-learning, huang-etal-2008-question}. Neural networks improve the classification performance with sentence embeddings~\cite{howard-ruder-2018-universal, cer-etal-2018-universal}. 

However, the aforementioned approaches in QA domain adaptation encourage domain-invariant features without considering samples from different classes and their distributional changes. Moreover, it is hitherto unclear how to estimate class discrepancies in QA, since class labels are not available in QA datasets. To solve this problem, we propose to use QC to divide QA pairs into different classes, where questions can be classified via an additional QC model or unsupervised clustering with minimum computational costs. We exploit the question classes by reducing the discrepancy among samples from the same class (‘intra-class’). Additionally, we design a distribution-aware sampling strategy in \ours to account for distributional changes between the source and target domains. By incorporating the discrepancy terms in the objective function, our self-supervised adaptation framework \ours achieves significant improvements against the state-of-the-art baseline methods.
\section{Methodology}
\label{sec:methodology}
\subsection{Setup}
\textbf{Data}: Our setting focuses on improving QA performance when domain shifts exist in the test data distribution. For this purpose, labeled source data and unlabeled target data are available, we denote the domain of source data with $\bm{\mathcal{D}}_{s}$ and target data with $\bm{\mathcal{D}}_{t}$. Formally, the input data is defined by:
\begin{itemize}
  \item \emph{Source data}: Labeled source data $\bm{X}_{s}$ from $\bm{\mathcal{D}}_{s}$. Individual sample $\bm{x}_{s}^{(i)} \in \bm{X}_{s}$ is defined by a triplet consisting of a question $\bm{x}_{s,q}^{(i)}$, a context $\bm{x}_{s,c}^{(i)}$, and an answer $\bm{x}_{s,a}^{(i)}$. The exact answer tokens can be found in context, answer $\bm{x}_{s,a}^{(i)}$ is represented by the start and end position in $\bm{x}_{s,c}^{(i)}$.
  \item \emph{Target data}: Unlabeled target data $\bm{X}_{t}$ from $\bm{\mathcal{D}}_{t}$. For target sample $\bm{x}_{t}^{(i)} \in \bm{X}_{t}$, we only have access to the question $\bm{x}_{t,q}^{(i)}$ and context $\bm{x}_{t,c}^{(i)}$. Ground truth answer $\bm{x}_{t,a}^{(i)}$ is not given for training.
\end{itemize}

\textbf{Model}: The QA system can be represented with function $\bm{f}$. $\bm{f}$ takes an input question $\bm{x}_{q}$ and context document $\bm{x}_{c}$ as input and yields answer prediction $\bm{x}_{a}$, namely $\bm{x}_{a} = \bm{f}(\bm{x}_{q}, \bm{x}_{c})$. The output $\bm{x}_{a}$ is represented as a subspan of $\bm{x}_{c}$ and comprises of the answer start and end positions.

\textbf{Objective}: The objective is to learn a $\bm{f}^{*}$, which maximizes the performance in answering questions from the target domain $\bm{\mathcal{D}}_{t}$. In other words, $\bm{f}^{*}$ minimizes the negative log likelihood (i.e., cross entropy) for $\bm{X}_{t}$ from the target domain distribution:
\begin{equation}
  \bm{f}^{*} = \arg \min_{\substack{\bm{f}}} \sum_{i=1}^{|\bm{X}_{t}|} \mathcal{L}_{\mathrm{NLL}}(\bm{f}(\bm{x}_{t,q}^{(i)}, \bm{x}_{t,c}^{(i)}), \bm{x}_{t,a}^{(i)}). \label{eq:objective}
\end{equation}

\subsection{The \ours Framework}
\subsubsection{Overall Framework}
In the proposed \ours, we design a self-supervised framework that facilitates question classification for QA domain adaptation. \ours can be divided into three stages: (1)~Question classification; (2)~Pseudo labeling \& sampling and (3)~Self-supervised adaptation. In the first stage, we perform classification for all input questions, which provides additional question class information for the adaptation stage. In the next stage, we label and filter all target samples and perform distribution-aware sampling to build mini-batches that resemble the target data distribution. Finally, we perform self-supervised adaptation on the QA system to minimize inter-domain and intra-class discrepancies. Once input questions are classified, we iteratively perform stage 2 and stage 3 in each epoch. The QA system is trained with both source and target data, where we encourage domain-invariant features and minimize intra-class discrepancies of data samples from the same question class.

Our approach leverages question classification for fine-grained domain adaptation. Here, QC is designed for evaluating intra-class discrepancies and distributional changes by introducing the additional question classes instead of using QA labels. The idea behind it is that QA labels are defined by subspans in input contexts, if we treat every combination of start and end position as a class, the corresponding label space would be too large and sparse for any meaningful discrepancy estimation. Therefore, we proposes the question classification stage to introduce additional semantic knowledge for intra-class discrepancy estimation. Moreover, by performing pseudo labeling and distribution-aware sampling, we resemble the target question distribution in the adaptation stage to correct the potential bias in the pretrained QA system. In other words, \ours simulates the target data distribution and `pulls together' source and target samples of the same question class to encourage domain invariance.

\subsubsection{Question Classification}
For question classification, we adopt the commonly used question taxonomy in TREC and categorize all questions into 6 coarse classes $Q$: \textbf{ABBR}: \emph{Abbreviation}, \textbf{DESC}: \emph{Description}, \textbf{ENTY}: \emph{Entity}, \textbf{HUM}: \emph{Human}, \textbf{LOC}: \emph{Location} and \textbf{NUM}: \emph{Numeric Value}. Each class indicates the potential answer type to the question~\cite{li-roth-2002-learning}. In practice, we rarely find \textbf{ABBR} questions.

\begin{figure}[t]
  \centering
  \includegraphics[width=1.0\linewidth]{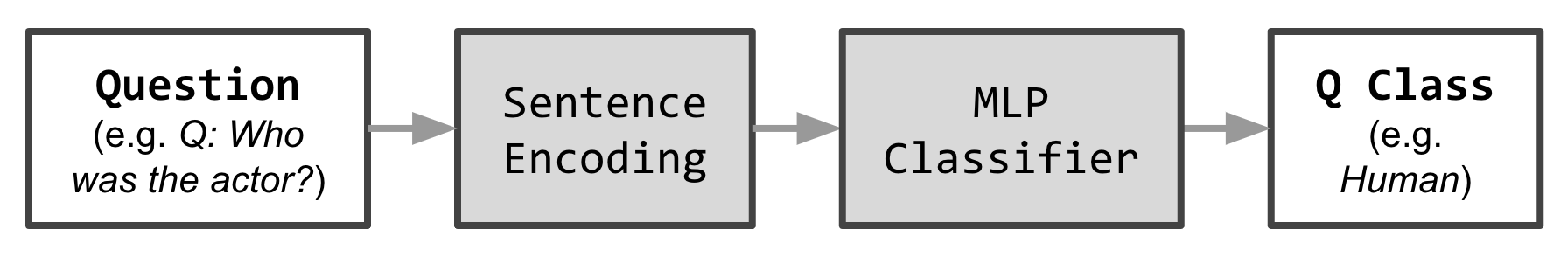}
  \caption{Model architecture for question classifier.}
  \label{fig:qc}
\end{figure}

The proposed QC model leverages pretrained sentence embedding methods to generate vectorized feature for input questions. We then build a multilayer perceptron (MLP) to perform classification on the encoded questions, see \Cref{fig:qc}. Specifically, we adopt InferSent and Universal Sentence Encoder to encode the input questions separately~\cite{conneau-etal-2017-supervised,cer-etal-2018-universal}. The encodings are concatenated and used as an input feature for the MLP classifier. With the trained QC model, inference can be performed on all training questions for the later adaptation stage. 

To further examine the effectiveness of question classification without additional model and annotation, we introduce an unsupervised clustering method, where we refrain from using an additional dataset or classifier to perform question classification. In particular, we feed the input data within the transformer encoder (part of the QA system) and utilize the output from the \texttt{[CLS]} token position as features~\cite{devlin-etal-2019-bert}. We sample a fixed number of source features (10k in our experiments) and perform KMeans clustering with a predefined number of clusters $k$ (Similar to TREC, we use 5 as default). Then, cluster centroids are preserved to classify source and target QA datasets.

\subsubsection{Pseudo Labeling \& Sampling}
Provided with the access to labeled source data, we pretrain the QA system $\bm{f}$ to learn to answer questions. After pretraining, we can use $\bm{f}$ to predict target answers for self-supervised adaptation. The pseudo labels are filtered according to the answer confidence, we preserve the target samples above confidence threshold $\lambda_{con}$. The pseudo labeling and confidence thresholding steps are repeated in each epoch to dynamically adjust the target distribution used for training.

For mini-batch training, we sample the same amounts of QA pairs from both domains to minimize the inter-domain and intra-class discrepancies. However, with randomly sampled data, training is less efficient as source and target questions in each batch can be entirely different (e.g., source samples are all \emph{Human} questions and target samples are all \emph{Description} questions). To solve this problem, we design a distribution-aware sampling strategy: we first sample target QA pairs from $\bm{X}_{t}$ and within the same question classes, we sample from $\bm{X}_{s}$ such that the source and target question classes in each batch are identical. Consequently, the QA system can be trained on a data distribution similar to the target dataset. Moreover, the estimation of intra-class discrepancy between both domains can be performed more efficiently.

\begin{figure*}[t]
  \centering
  \includegraphics[width=1.0\linewidth]{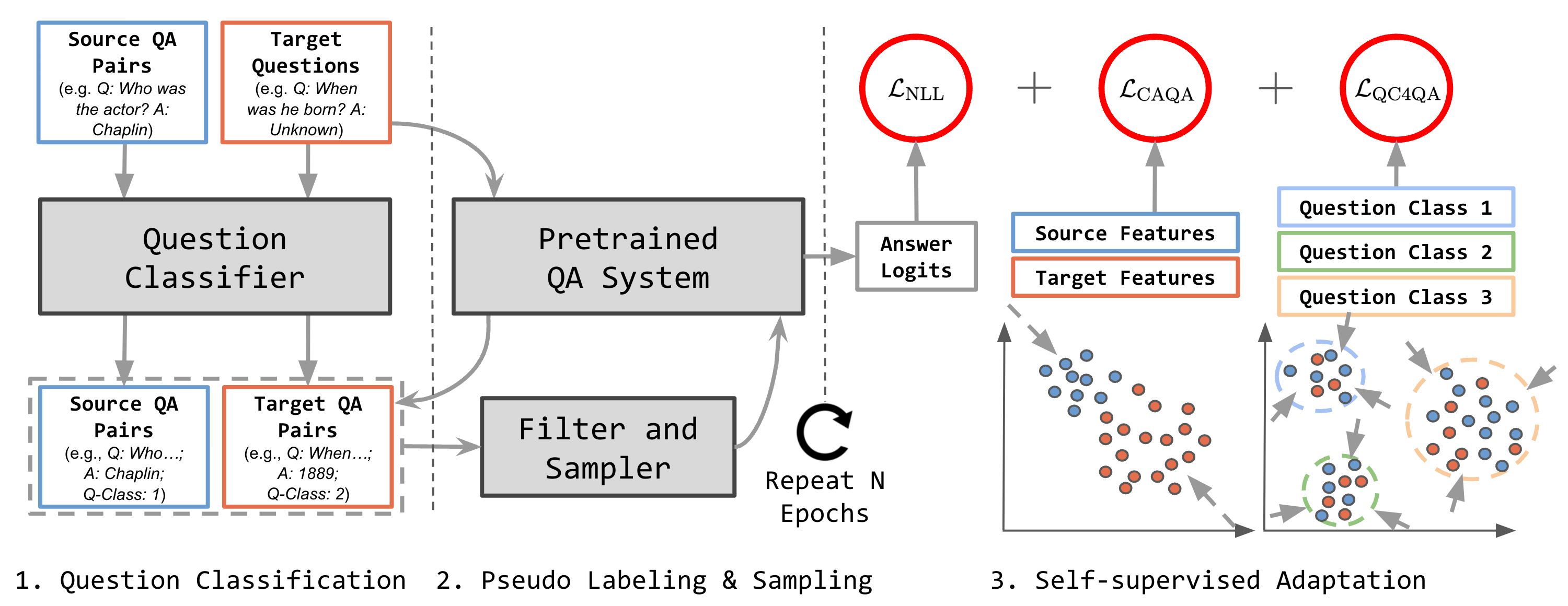}
  \caption{Overview for the proposed method. \ours can be divided into three stages: (1)~Question classification where all questions are assigned different classes; (2)~Pseudo labeling \& sampling, where we label and sample target examples with the proposed distribution-aware sampling strategy; (3)~Self-supervised adaptation, in which we train the QA system jointly with source and target data. In the experiments, stage 2 and 3 are iteratively performed and we apply the proposed objective \Cref{eq:overall-objective} to minimize both inter-domain and intra-class discrepancies.}
  \label{fig:method}
\end{figure*}

\subsubsection{Self-supervised Adaptation}
The sampled batches are used to adapt the pretrained QA system, where we optimize the model to reduce the negative log likelihood loss, as in \Cref{eq:objective}. Meanwhile, we encourage the domain invariance by computing the discrepancies and incorporate them in the training objective.

To measure the discrepancy between samples from different domains, we adopt the maximum mean discrepancy (MMD) distance~\cite{gretton2012kernel}. MMD estimates the distance between two distributions with samples drawn from them, with $f$ and $\mathcal{H}$ representing the feature mapping and the reproducing kernel Hilbert space:
\begin{equation}
  \mathcal{D} = \sup_{f \in \mathcal{H}} \big( \frac{1}{|\bm{X}_{s}|} \sum_{i=1}^{|\bm{X}_{s}|}f(\bm{x}_{s}^{(i)}) - \frac{1}{|\bm{X}_{t}|} \sum_{i=1}^{|\bm{X}_{t}|}f(\bm{x}_{t}^{(i)}) \big). 
  \label{eq:mmd}
\end{equation}

To simplify the computation, we adopt the Gaussian kernel as feature mapping, i.e., $k(\bm{x}_{s}^{(i)}, \bm{x}_{t}^{(j)}) = \mathrm{exp}(- \frac{\| \bm{x}_{s}^{(i)} - \bm{x}_{t}^{(j)} \|^{2}}{\gamma})$. We further leverage the kernel trick and empirical kernel mean embeddings~\cite{long2015learning} to estimate the squared MMD distance between samples from $\bm{X}_{s}$ and $\bm{X}_{t}$:
\begin{equation}
  \begin{aligned}
    \mathcal{D}^{\mathrm{MMD}} &= \frac{1}{|\bm{X}_{s}||\bm{X}_{s}|} \sum_{i=1}^{|\bm{X}_{s}|} \sum_{j=1}^{|\bm{X}_{s}|} k(\phi(\bm{x}_\mathrm{s}^{(i)}), \phi(\bm{x}_\mathrm{s}^{(j)})) \\
    &+ \frac{1}{|\bm{X}_{t}||\bm{X}_{t}|} \sum_{i=1}^{|\bm{X}_{t}|} \sum_{j=1}^{|\bm{X}_{t}|} k(\phi(\bm{x}_\mathrm{t}^{(i)}), \phi(\bm{x}_\mathrm{t}^{(j)})) \\
    &- \frac{2}{|\bm{X}_{s}||\bm{X}_{t}|} \sum_{i=1}^{|\bm{X}_{s}|} \sum_{j=1}^{|\bm{X}_{t}|} k(\phi(\bm{x}_\mathrm{s}^{(i)}), \phi(\bm{x}_\mathrm{t}^{(j)})),
  \label{eq:contrastive_loss}
  \end{aligned}
\end{equation}
where $\phi$ represents the transformer encoder in the QA system. With $\mathcal{D}^{\mathrm{MMD}}$, it is possible to measure the discrepancies between different domains and question classes. The discrepancy values are used to guide the self-supervised adaptation and encourage domain-invariant features.

Among all tokens in each QA sample $\bm{x}$, we distinguish two types of features $\bm{x}_\mathrm{a}$ and $\bm{x}_\mathrm{o}$. $\bm{x}_\mathrm{a}$ stands for the mean vector of answer token representations, while $\bm{x}_\mathrm{o}$ is the mean vector of all other tokens in the representation space~\cite{yue-etal-2021-contrastive}. The QA system extracts the answer when the answer tokens in the representation space are separated from $\bm{x}_\mathrm{o}$~\cite{van2019does}. Therefore, we adopt $\bm{x}_\mathrm{a}$ as feature representation to compute MMD distances. By introducing question classification, we introduce an additional term w.r.t. the intra-class discrepancy in the objective function: 
\begin{equation}
  \begin{aligned}
    \mathcal{L}_{\text{\ours}} = \frac{1}{|Q|}\sum_{q \in Q}^{Q}\mathcal{D}^{\mathrm{MMD}}(\bm{X}_{s}^{(q)}, \bm{X}_{t}^{(q)}),
  \label{eq:ours_loss}
  \end{aligned}
\end{equation}
$\bm{X}_{s}$ refers to all answer features in source samples and $\bm{X}_{t}$ represents answer features in target samples,
while $\bm{X}^{(q)}$ denotes the set of samples that belong to question class $q \in Q$. $\mathcal{L}_{\text{\ours}}$ `pulls together' features from the same question class across domains by minimizing their MMD distances.

\subsubsection{Overall Objective}

To encourage domain-invariant features, we incorporate \Cref{eq:ours_loss} into the training objective. Using the NLL loss and the contrastive adaptaion loss~\cite{yue-etal-2021-contrastive, yue2022contrastive}, the overall objective function can be formulated as follows:
\begin{equation}
  \mathcal{L} = \mathcal{L}_{\mathrm{NLL}} + \lambda (\mathcal{L}_{\mathrm{CAQA}} + \mathcal{L}_{\mathrm{\ours}}),
  \label{eq:overall-objective}
\end{equation}
in which $\mathcal{L}_{\mathrm{CAQA}}$ is the same as in~\cite{yue-etal-2021-contrastive}, while $\lambda$ is a scaling factor we choose empirically. Although we introduce $\mathcal{L}_{\mathrm{CAQA}}$ in our training objective function, \ours is largely different from CAQA as: (1)~$\mathcal{L}_{\mathrm{CAQA}}$ only reduces the inter-domain discrepancy, we incorporate question classification to additionally reduce intra-class discrepancy via $\mathcal{L}_{\mathrm{\ours}}$ for fine-grained adaptation; (2)~We perform pseudo labeling and distribution-aware sampling to account for the distribution shifts between the source and target dataset; and (3)~\ours leverages an efficient self-supervised adaptation framework instead of the computationally expensive question generation in~\cite{yue-etal-2021-contrastive}. As such, the proposed \ours efficiently reduces the domain discrepancy and effectively transfers learnt knowledge from the source domain to the target domain.

The overall framework is illustrated in \Cref{fig:method}. We first generate question classes for all data samples. Next, the source-pretrained QA model generates pseudo labels for target data and we select target samples above the confidence threshold $\lambda_{con}$ for training. Pseudo labeling and self-supervised adaptation are performed iteratively to refine the pseudo labels and improve the performance in the target domain using \Cref{eq:overall-objective}. Unlike previous works~\cite{lee-etal-2019-domain, cao2020unsupervised, shakeri-etal-2020-end, yue-etal-2021-contrastive}, we discard domain-adversarial training or question generation and introduce a self-supervised adaptation framework based on question classification for improved efficiency and adaptation performance. We also design a distribution-aware sampling strategy to resemble the target data distribution and correct the potential bias in the pretrained QA system. Additionally, a fine-grained adaptation loss based on question classification is introduced in training to minimize both the inter-domain and intra-class discrepancies across the source and target domain.

\begin{table*}[t]
\centering
\begin{tabular}{lccccc}
\toprule
\multirow{2}{*}{\textbf{Model}}                          & \textbf{CNN}            & \textbf{Daily Mail}     & \textbf{NewsQA}         & \textbf{HotpotQA}       & \textbf{SearchQA}       \\
                                                         & EM / F1                 & EM / F1                 & EM / F1                 & EM / F1                 & EM / F1                 \\ \midrule
\multicolumn{6}{c}{(I) Zero-shot target performance}                                                                                                                                       \\ \midrule
BERT-QA                                                  & 14.30/23.57             & 15.38/25.90             & 39.17/56.14             & 43.34/60.42             & 16.19/25.03             \\ \midrule
\multicolumn{6}{c}{(II) Target performance with domain adaptation}                                                                                                                         \\ \midrule
DAT~\cite{lee-etal-2019-domain}                          & 21.89/27.37             & 26.98/32.72             & 38.73/54.24             & 44.25/61.10             & 22.31/31.64             \\
CASe~\cite{cao2020unsupervised}                          & 20.77/29.37             & 25.40/35.85             & 43.43/59.67             & 47.16/63.88             & 26.07/35.16             \\
CAQA\textsuperscript{*}~\cite{yue-etal-2021-contrastive} & 21.97/30.97             & 32.08/41.47             & 44.26/60.83             & 48.52/64.76             & 32.05/41.07             \\
\ours KMeans (Ours)                                      & \underline{25.04/33.20} & \underline{35.53/44.32} & \underline{44.40/60.91} & \underline{49.58/65.78} & \underline{34.44/43.78} \\
\ours TREC (Ours)                                        & \textbf{28.05/36.18}    & \textbf{36.43/45.85}    & \textbf{45.62/61.71}    & \textbf{50.02/66.10}    & \textbf{35.75/44.37}    \\ \bottomrule
\end{tabular}
\caption[Main results]{Main results of QA adaptation performance on target dataset.}
\label{tab:main-results}
\end{table*}

\begin{table*}[t]
\centering
\begin{tabular}{lcccc}
\toprule
\multirow{2}{*}{\textbf{Model}}                          & \textbf{CoQA}           & \textbf{DROP}           & \textbf{Natural Questions} & \textbf{TriviaQA}       \\
                                                         & EM / F1                 & EM / F1                 & EM / F1                 & EM / F1                 \\ \midrule
\multicolumn{5}{c}{(I) Zero-shot target performance}                                                                                                             \\ \midrule
BERT-QA                                                  & 12.42/17.30             & 19.36/30.28             & 39.06/53.75             & 49.70/59.09             \\ \midrule
\multicolumn{5}{c}{(II) Target performance with domain adaptation}                                                                                               \\ \midrule
DAT~\cite{lee-etal-2019-domain}                          & 11.98/14.72             & 18.53/29.34             & 44.94/58.91             & 49.94/59.82             \\
CASe~\cite{cao2020unsupervised}                          & 13.71/18.57             & 21.78/31.44             & 46.53/60.19             & 54.74/63.61             \\
CAQA\textsuperscript{*}~\cite{yue-etal-2021-contrastive} & 14.41/19.28             & 22.48/31.56             & 47.37/60.52             & 54.30/62.98             \\
\ours KMeans (Ours)                                      & \underline{14.83/19.60} & \underline{23.13/31.73} & \underline{49.37/62.25} & \underline{54.99/63.58} \\
\ours TREC (Ours)                                        & \textbf{15.03/19.71}    & \textbf{23.46/32.22}    & \textbf{50.59/62.98}    & \textbf{55.98/64.57}    \\ \bottomrule
\end{tabular}
\caption[Additional results]{Results of QA adaptation performance on additional target dataset.}
\label{tab:additional-results}
\end{table*}
\section{Experiments}
\label{sec:experiments}

\subsection{Datasets and Baselines}
For supervised question classification, we adopt the TREC dataset~\cite{li-roth-2002-learning}, a widely used dataset containing \textasciitilde5k training questions and 500 questions for testing. Following~\cite{cao2020unsupervised, shakeri-etal-2020-end, yue-etal-2021-contrastive}, we use SQuAD as our source domain QA dataset~\cite{rajpurkar-etal-2016-squad}. For target domain, we adopt multiple QA datasets (details in \Cref{sec:dataset-baseline}) and refrain from using labels in training~\cite{cao2020unsupervised, shakeri-etal-2020-end, yue-etal-2021-contrastive}.

For comparison, we adopt 4 baseline methods. We first pretrain a QA system on the source dataset and then evaluate on each target dataset with zero knowledge of the target domain. We additionally adopt 3 state-of-the-art baselines: Domain-adversarial training (DAT)~\cite{lee-etal-2019-domain}, conditional adversarial self-training (CASe)~\cite{cao2020unsupervised} and contrastive adaptation for QA (CAQA\textsuperscript{*})~\cite{yue-etal-2021-contrastive}. For fair comparison, we adapt the original CAQA to our self-supervised adaptation framework as a baseline, we denote the adapted CAQA with CAQA\textsuperscript{*}.\footnote{We exclude question generation and adopt the same process of pseudo labeling, distribution-aware sampling and self-supervised adaptation as \ours in CAQA\textsuperscript{*}. Different from \ours, we use the same objective function as in~\cite{yue-etal-2021-contrastive}. A direct comparison between the proposed \ours and the original CAQA can be found in \Cref{sec:caqa-comparison}.} BERT-QA is selected as the QA model~\cite{devlin-etal-2019-bert}. Details of the baselines are elaborated in \Cref{sec:dataset-baseline}.

\subsection{Training and Evaluation}
We train our QC model on the TREC training set and evaluate on the test set,
the best model is saved to perform classification on all QA datasets. For unsupervised question classification, sampled \texttt{[CLS]} features from the source dataset are used to perform KMeans clustering, followed by question class inference on all QA datasets.

After question classification, we adopt a QA model (pretrained on the source dataset) and iteratively perform: (1)~Pseudo labeling and distribution-aware sampling to select data batches that resemble the target data distribution; (2)~Self-supervised adaptation with the proposed objective \Cref{eq:overall-objective} for learning domain-invariant representation. For evaluation, we adopt two metrics: exact match (EM) and F1 score (F1). We compute the metrics on target dev sets to evaluate the adaptation performance. Details of our implementation can be found in \Cref{sec:implementation}.

\subsection{Main Results}
We first report the question classification performance on TREC dataset. The MLP classifier has 2.36M parameters and can be trained efficiently in less than one minute (57.6s on average) with GPU acceleration. We perform the evaluation with the proposed MLP QC model and reach an accuracy of 96.6\% on the TREC test set. Similar magnitude of efficiency can be observed in KMeans clustering for unsupervised question classification. We provide detailed quantitative analysis and qualitative examples in \Cref{sec:qc-results}.

The QA system is first pretrained in the source domain with 79.60 EM and 87.64 F1 score on the SQuAD dev set. Then, we perform adaptation experiments and report the main results in \Cref{tab:main-results}, results on the additional target datasets can be found in \Cref{tab:additional-results}. Both tables are divided into 2 parts: (1)~QA systems pretrained on SQuAD as na{\"i}ve baseline (`Zero-shot target performance'); (2)~Baseline methods and \ours for QA domain adaptation (`Target performance with domain adaptation'). The proposed approach with TREC supervised classification is denoted with `\ours TREC', unsupervised KMeans question clustering is denoted with `\ours KMeans'.

The following observations can be made from our experiments: (1)~Unsupervised adaptation methods achieve superior performance than the na{\"i}ve baseline in most cases. Compared to the na{\"i}ve baseline, \ours can lead to improvements of over 100\% in EM and and over 75\% in F1. (2)~Compared to contrastive adaptation (e.g., CAQA\textsuperscript{*}), the proposed \ours is particularly effective on cloze questions (i.e., CNN and Daily Mail), with average performance gains of 16.5\% and 10.4\% in EM and F1. This suggests that we can benefit more from QC when the target questions are less similar to source questions. (3)~By comparing CAQA\textsuperscript{*} and both \ours methods, we find consistent performance improvements due to question classification for all datasets. (4)~Both \ours methods outperform baseline methods with considerable improvements, from which \ours TREC demontrates the best performance on all datasets. For example, \ours KMeans significantly outperforms the best baseline CAQA\textsuperscript{*} with 5.2\% and 3.2\% performance increases in EM and F1 on average. For \ours TREC, the relative improvements are 8.6\% and 5.5\% respectively. Altogether, the results suggest that both the TREC and KMeans question classification are effective for improving the performance on out-of-domain data. Additional results and analysis can be found in \Cref{sec:additional-results}.

\subsection{Ablation Studies}
\subsubsection{Question Classification}
We first study the benefits of question classification. The performance gains can be achieved by comparing the results between CAQA\textsuperscript{*} and \ours in \Cref{tab:main-results} and \Cref{tab:additional-results}. This is because CAQA\textsuperscript{*} is adapted to the same self-supervised adaptation framework as in \ours. CAQA\textsuperscript{*} models are trained without minimizing the intra-class discrepancies in \Cref{eq:overall-objective}. The improvements from \ours suggest that both supervised and unsupervised question classification can consistently improve QA systems in answering questions from unseen domains. Moreover, \ours is particularly effective on target datasets with different question formats (e.g., CNN and Daily Mail).

\subsubsection{Distribution-aware Sampling}
To study the influence of distribution-aware sampling in \ours, we replace the distribution-aware sampling strategy with random sampling. Then we perform unsupervised adaptation with \ours TREC on CNN, Daily Mail and NewsQA to verify the merits of the sampling strategy. Results are presented in \Cref{tab:sampling}.

\begin{table}[h]
  \centering
  \begin{tabular}{lccc}
  \toprule
  \multirow{2}{*}{Dataset} & \textbf{Rand. Sampling} & \textbf{\ours} \\
                           & EM / F1                 & EM / F1        \\ \midrule
  CNN        & 26.69/35.24 & \textbf{28.05/36.18} \\
  Daily Mail & 35.83/45.51 & \textbf{36.43/45.85} \\
  NewsQA     & 44.72/61.02 & \textbf{44.86/61.40} \\ \bottomrule
  \end{tabular}
  \caption[\ours performance w/ random sampling]{\ours performance with the random sampling strategy.}
  \label{tab:sampling}
\end{table}

In all target datasets, we see performance drops when we replace the proposed strategy with random sampling. In particular, we find relatively large performance deterioration on CNN without distribution-aware sampling. We believe the reason is that the question distribution in CNN is less similar to SQuAD (see \Cref{tab:qc-distribution}), the resulting inconsistency in sampled batches reduces the effectiveness in discrepancy estimation.

\subsubsection{Sensitivity of Hyperparameter $\lambda$}
Now we evaluate the influence of $\lambda$ to study the robustness of the proposed objective function. We select different values ranging from 0 to 5e-2 and perform adaptation with \ours TREC. Experiments on CNN, Daily Mail and NewsQA are presented to estimate the influence of $\lambda$.

\begin{figure}[h]
  \centering
  \includegraphics[width=1.0\linewidth]{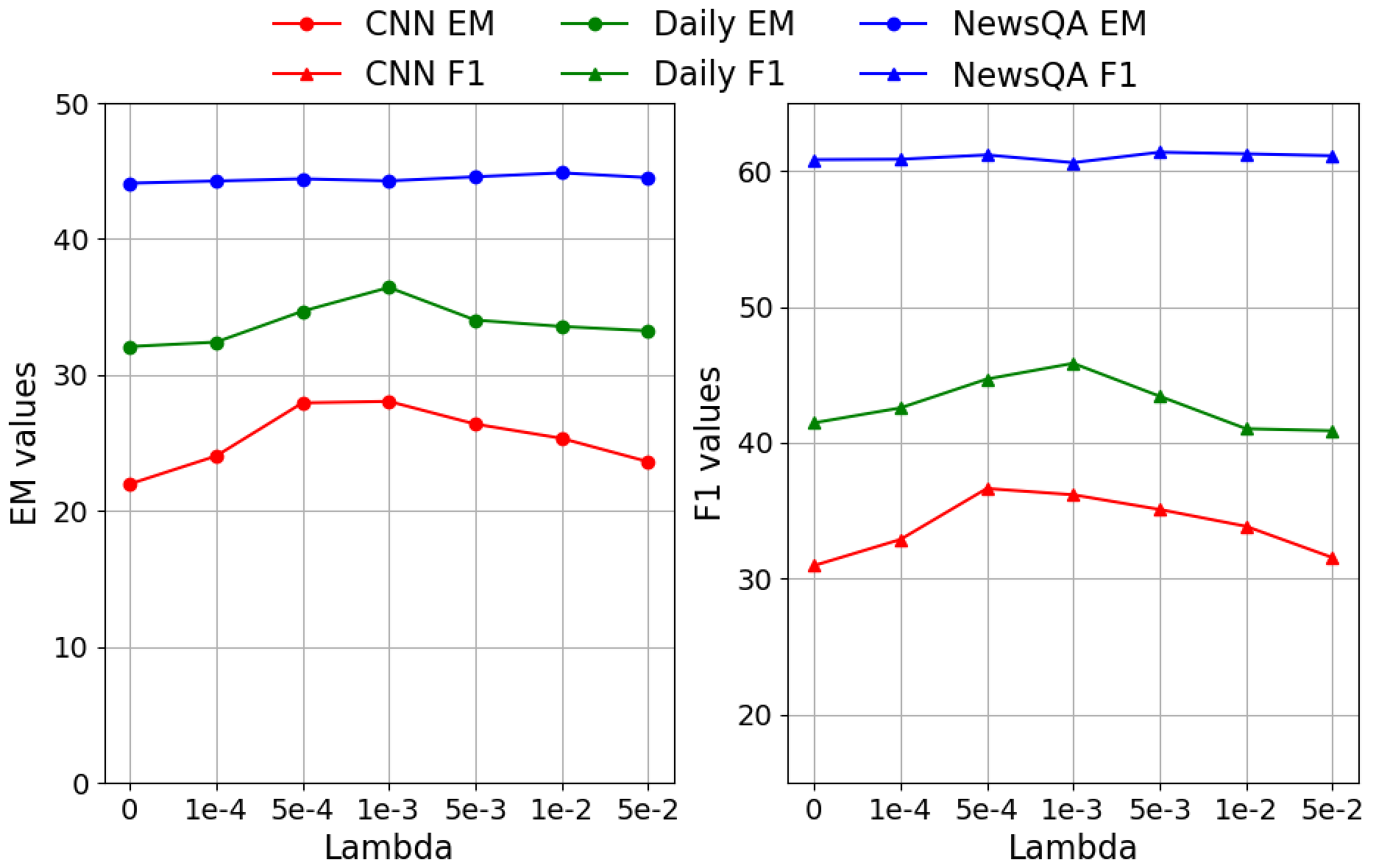}
  \caption{\ours adaptation performance for different lambda values. X-axis represents lambda and y-axis represents EM / F1 scores.}
  \label{fig:lambda}
\end{figure}

\Cref{fig:lambda} visualizes EM / F1 with increasing $\lambda$. Despite certain variations, we observe the results first go up and then decrease. Although CNN and Daily are more sensitive to $\lambda$, we observe greater improvements by reducing inter-domain and intra-class discrepancies. Overall, \ours consistently improves adaptation performance.

\subsubsection{Confidence Threshold in Pseudo labeling}
We study the influence of $\lambda_{con}$ to understand how the performance varies with different confidence thresholds in pseudo labeling. We select different threshold values ranging from 0.2 to 0.8 to filter pseudo labels and train with \ours TREC. Experiments are performed on HotpotQA and SearchQA to estimate the influence of $\lambda_{con}$.

\begin{table}[h]
  \centering
  \begin{tabular}{cccc}
  \toprule
  \multirow{2}{*}{$\lambda_{con}$ Selection} & \textbf{HotpotQA} & \textbf{SearchQA} \\
                                   & EM / F1                 & EM / F1           \\ \midrule
  0.2      & 48.12/64.56          & 25.60/34.50 \\
  0.4      & \textbf{50.02/66.10} & 33.56/42.62 \\
  0.6      & 49.95/69.84          & 35.75/44.37 \\ 
  0.8      & 49.44/65.11          & \textbf{37.29/46.31} \\ \bottomrule
  \end{tabular}
  \caption[\ours performance w/ random sampling]{QC4QA adaptation performance for different confidence thresholds in pseudo labeling.}
  \label{tab:con-thres}
\end{table}

\Cref{tab:con-thres} shows adaptation performance with different $\lambda_{con}$. The best performance can be reached with $\lambda_{con}$ ranging from 0.4 to 0.8. For large datasets like SearchQA (with over 100k QA pairs), a higher confidence threshold yields a better adaptation performance since we avoid noisy pseudo labels. In sum, carefully selected $\lambda_{con}$ yields comparatively large improvements for \ours.

\section{Conclusion}
\label{sec:conclusion}

In this paper, we propose a novel framework for QA domain adaptation. The proposed \ours combines question classification with self-supervised adaptation techniques. \ours leverages question classes to reduce domain discrepancies and resemble target data distribution in training. Different from existing works, \ours achieves superior performance by introducing a simple question classifier and incorporating the question class information in the training objective. We demonstrate the efficiency and effectiveness of \ours compared to state-of-the-art approaches by achieving a substantially better performance on multiple datasets.

Despite having adopted question classification to adapt QA systems to unseen target domains, the proposed \ours has certain limitations. For example, we assume access to unlabeled questions in QA datasets and have not exploited the potential benefits of different question samples and question classes. For future work, we plan to relax our settings and explore question generation and question value estimation for QA domain adaptation.

\section*{Acknowledgments}

This research is supported in part by the National Science Foundation under Grant No. IIS-2202481, CHE-2105005, IIS-2008228, CNS-1845639, CNS-1831669. The views and conclusions contained in this document are those of the authors and should not be interpreted as representing the official policies, either expressed or implied, of the U.S. Government. The U.S. Government is authorized to reproduce and distribute reprints for Government purposes notwithstanding any copyright notation here on.

\bibliography{anthology, reference}
\bibliographystyle{acl_natbib}

\clearpage
\appendix
\section{Dataset and baseline details}
\label{sec:dataset-baseline}

\subsection{Dataset details}
For QA datasets, we follow~\cite{lee-etal-2019-domain, cao2020unsupervised, yue-etal-2021-contrastive} and select SQuAD v1.1 as our source dataset~\cite{rajpurkar-etal-2016-squad}. SQuAD is a crowdsourced QA dataset based on Wikipedia articles. For target domain, we adopt multiple datasets to evaluate \ours:
\begin{enumerate}
\item \textbf{CNN}~\cite{hermann2015teaching} leverages CNN articles as contexts. Cloze QA pairs are generated by replacing answers with `@placeholder'.
\item \textbf{CoQA}~\cite{reddy-etal-2019-coqa} is a conversational dataset with rationales and QA pairs. Contexts are given as multi-turn conversations.
\item \textbf{Daily Mail}~\cite{hermann2015teaching} is similar to CNN and consists of news from Daily Mail. Cloze questions and answers are used.
\item \textbf{DROP}~\cite{dua2019drop} requires QA systems to resolve references, reasoning, matching and understanding context implications.
\item \textbf{NewsQA}~\cite{trischler2016newsqa} provides news as contexts and challenging questions beyond simple matching and entailment.
\item \textbf{HotpotQA}~\cite{yang-etal-2018-hotpotqa} provides multi-hop questions with challenging contexts (distractor contexts excluded).
\item \textbf{Natural Questions}~\cite{kwiatkowski-etal-2019-natural} has user questions. We adopt short answers and use long answers as contexts.
\item \textbf{TriviaQA}~\cite{joshi-etal-2017-triviaqa} is a large-scale QA dataset that includes QA pairs and supporting facts for supervised training.
\item \textbf{SearchQA}~\cite{dunn2017searchqa} is constructed through existing QA pairs by searching for context from online search results.
\end{enumerate}

\subsection{Baseline details}
For na{\"i}ve baseline, we adopt BERT-QA (uncased base version with additional batch normalization layer) and train on the source dataset~\cite{devlin-etal-2019-bert, cao2020unsupervised}. 
Additionally, we select 3 baselines in unsupervised QA domain adaptation:
\begin{enumerate}
\item \textbf{Domain adversarial training (DAT)}~\cite{lee-etal-2019-domain} comprises of a QA system and a discriminator using \texttt{[CLS]} output in BERT. The QA system is first trained on labeled source data. Then, input data from both domains is used for domain-adversarial training to learn generalized features.
\item \textbf{Conditional adversarial self-training (CASe)}~\cite{cao2020unsupervised} leverages self-training with domain-adversarial learning. CASe iteratively perform self-training and domain adversarial training to reduce domain discrepancy. We adopt the entropy weighted version CASe+E in our work as baseline. 
\item \textbf{Contrastive adaptation for QA (CAQA\textsuperscript{*})}~\cite{yue-etal-2021-contrastive} proposes contrastive adaptation based on token-level features. CAQA utilizes answer tokens as features and reduce the domain gap by minimizing MMD distances. We exclude question generation and adopt the same process of pseudo labeling, distribution-aware sampling and self-supervised adaptation. In particular, we perform training using the original contrastive adaptation loss as in~\cite{yue-etal-2021-contrastive}.
\end{enumerate}

\section{Implementation}
\label{sec:implementation}
We first train a question classifier on the TREC dataset. The QC model is trained for 4 epochs using RMSprop optimizer with learning rate of 0.01 and batch size of 64. We evaluate the QC model on the TREC test set and report the accuracy of the best QC model.

For pretraining BERT-QA on the source dataset (i.e., SQuAD), we follow~\cite{devlin-etal-2019-bert, yue-etal-2021-contrastive} to preprocess data and perform training. We select the AdamW optimizer and train BERT-QA for 2 epochs without linear warmup. Learning rate is 3e-5 and batch size is 12. After pretraining, we validate the model with the provided dev set and report the EM and F1 scores.

For baseline methods, we use our pretrained BERT-QA and follow their default settings for domain adaptation. For \ours, adaptation is performed 4 epochs with the AdamW optimizer, learning rate of 3e-5 and 10\% proportion as warmup in training. In the pseudo labeling stage, we first perform inference on unlabeled target data and preserve the target samples above confidence threshold $\lambda_{con}$. For batching in self-supervised adaptation stage, we sample 12 target examples and perform distribution-aware sampling to sample another 12 source QA pairs. The sampled source data has same question classes as the target examples. Validation is performed every 2000 iterations and after every epoch to save the best QA model. In our experiments, we empirically select $\lambda$ from [1e-4, 1e-3, 1e-2], we select $\lambda_{con}$ from [0.4, 0.6]. Our system setup is Intel Xeon Gold 6326 CPU, NVIDIA A40 GPU and 128GB RAM.

\section{Additional results}
\label{sec:additional-results}

\begin{table*}[t]
\centering
\begin{tabular}{lcccccc}
\hline
\textbf{Dataset}           & \textbf{ABBR}   & \textbf{DESC}   & \textbf{ENTY}   & \textbf{HUM}   & \textbf{LOC}   & \textbf{NUM}   \\ \hline
\textbf{SQuAD}             & 0.5\%           & 12.5\%          & 31.4\%          & 19.9\%         & 11.6\%         & 24.1\%         \\
\textbf{CNN}               & 0.0\%           & 5.3\%           & 39.2\%          & 43.6\%         & 5.3\%          & 6.5\%          \\
\textbf{Daily Mail}        & 0.0\%           & 3.9\%           & 38.6\%          & 46.1\%         & 4.4\%          & 6.9\%          \\
\textbf{NewsQA}            & 0.1\%           & 19.5\%          & 21.4\%          & 29.4\%         & 10.3\%         & 19.3\%         \\
\textbf{HotpotQA}          & 0.0\%           & 1.6\%           & 21.1\%          & 51.6\%         & 14.8\%         & 10.9\%         \\
\textbf{CoQA}              & 0.1\%           & 14.1\%          & 14.2\%          & 53.7\%         & 9.3\%          & 8.5\%          \\
\textbf{DROP}              & 0.0\%           & 2.3\%           & 24.0\%          & 51.8\%         & 7.3\%          & 14.5\%         \\
\textbf{Natural Questions} & 0.2\%           & 5.0\%           & 13.2\%          & 39.2\%         & 13.4\%         & 29.0\%         \\
\textbf{SearchQA}          & 0.2\%           & 3.7\%           & 43.1\%          & 4.8\%          & 13.5\%         & 34.7\%         \\
\textbf{TriviaQA}          & 0.2\%           & 1.6\%           & 37.8\%          & 36.5\%         & 19.3\%         & 4.7\%          \\ \hline
\end{tabular}
\caption[Question classification results]{Question class distribution in all datasets.}
\label{tab:qc-distribution}
\end{table*}

\begin{table*}[t]
\centering
\begin{tabular}{lcccc}
\hline
\multirow{2}{*}{\textbf{Model}}                          & \textbf{Natural Questions} & \textbf{HotpotQA}       & \textbf{SearchQA}                & \textbf{TriviaQA}       \\
                                                         & EM / F1                    & EM / F1                 & EM / F1                          & EM / F1                 \\ \hline
\multicolumn{5}{c}{(I) Zero-shot target performance}                                                                                                                         \\ \hline
BERT-QA                                                  & 39.06/53.75                & 43.34/60.42             & 16.19/25.03                      & 49.70/59.09             \\ \hline
\multicolumn{5}{c}{(II) Target performance with domain adaptation}                                                                                                           \\ \hline
CAQA\textsuperscript{*}~\cite{yue-etal-2021-contrastive} & 47.37/60.52                & 48.52/64.76             & 32.05/41.07                      & 54.30/62.98             \\
CAQA~\cite{yue-etal-2021-contrastive}                    & 48.55/\underline{62.60}    & 46.37/61.57             & \textbf{36.05}/42.94             & \underline{55.17}/63.23 \\
\ours KMeans (Ours)                                      & \underline{49.37}/62.25    & \underline{49.58/65.78} & 34.44/\underline{43.78}          & 54.99/\underline{63.58} \\
\ours TREC (Ours)                                        & \textbf{50.59/62.98}       & \textbf{50.02/66.10}    & \underline{35.75}/\textbf{44.37} & \textbf{55.98/64.57}    \\ \hline
\end{tabular}
\caption[Comparison between \ours and CAQA]{Comparison between \ours and CAQA.}
\label{tab:caqa-comparison}
\end{table*}

\subsection{Question Classification Results}
\label{sec:qc-results}
Due to our light-weight design, the TREC question classifier can perform training and inference efficiently within a few minutes. For example, we achieve an average training time of 57.6s on TREC in repeated experiments with GPU acceleration. Inference on QA datasets are of similar efficiency and depends on the individual size of each dataset.

Since TREC classes are not provided in QA datasets, it's not possible to directly evaluate the supervised QC model on them. We report the distribution of different question classes in \Cref{tab:qc-distribution}, where we observe significant distribution shifts between the source dataset and certain target datasets (e.g., CNN and Daily Mail). Additionally, we present selected examples of classified questions in \Cref{tab:qc-examples}, from which we observe the following: (1)~Cloze questions are more difficult to classify. Unlike natural questions, cloze questions usually do not contain auxiliary verb and wh-words (e.g., what, where etc.) as indicator of the question classes. (2)~Multiple question classes may qualify for cloze questions. In some examples, different types of tokens can be filled in the placeholder position (e.g., both \textbf{DESC} and \textbf{ENTY} qualify for Q3). (3)~The TREC question classifier can be less accurate on cloze questions. This is the case for Q5 in \Cref{tab:qc-examples}, where the questions are more likely to be \textbf{DESC} and \textbf{LOC} than \textbf{HUM}. (4)~For natural questions, the question classifier performs generally well and makes fewer mistakes due to the similarity of natural questions across QA datasets. More examples can be found in the released code and data.

\begin{figure}[h]
  \centering
  \includegraphics[width=1.0\linewidth]{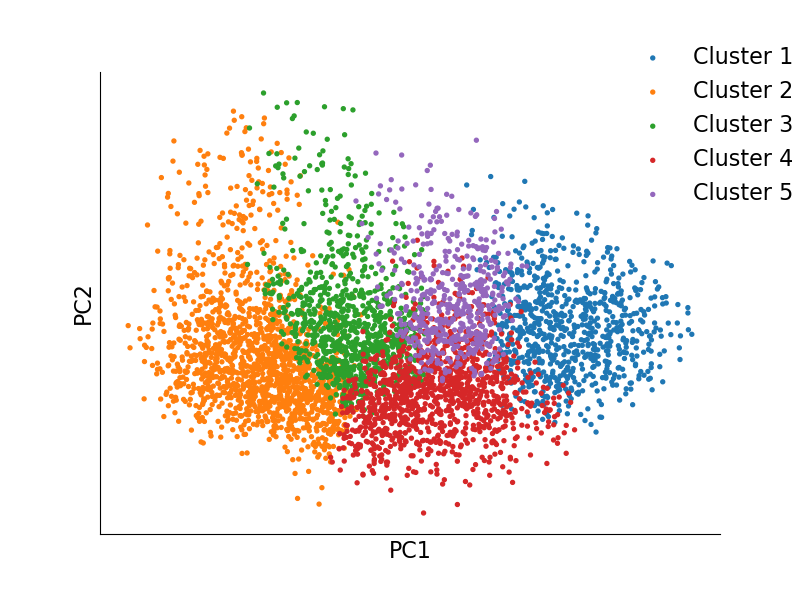}
  \caption{Visualization of KMeans cluster analysis on NewsQA using \texttt{[CLS]} features from BERT.}
  \label{fig:cluster}
\end{figure}

For KMeans unsupervised question classification, we focus on the discrepancy among question samples and perform KMeans clustering using the \texttt{[CLS]} output from BERT encoder, see \Cref{fig:cluster}. The plot shows a principle component analysis (PCA) visualizing the BERT-encoder output of NewsQA examples, where different colors represent question class predictions via KMeans algorithm. We observe that \texttt{[CLS]} features are comparatively homogeneous, making it hard to determine cluster boundaries that clearly separate different classes of questions. This might cause performance deterioration in case of increasing outliers. Overall, KMeans can successfully cluster QA examples within each neighborhood on the target dataset. Ideally, the cluster labels can be used to reduce intra-class discrepancies for fine-grained domain adaptation similar to TREC classification. Both adaptation results and cluster visualization suggest that KMeans is effective in improving the performance on out-of-domain data.

\subsection{Comparison with CAQA}
\label{sec:caqa-comparison}

We study the effectiveness of the proposed \ours by comparing the performance between \ours, the original CAQA and the adapted CAQA\textsuperscript{*}~\cite{yue-etal-2021-contrastive}. The results are presented in \Cref{tab:caqa-comparison}, we observe that the best-performing method is \ours with TREC question classification for 7 out of 8 metric values. For SearchQA, the original CAQA performs the best in EM, with \ours TREC is of similar magnitude and clearly ranks second. On average, \ours TREC performs the best with EM of 48.09 and F1 of 59.51. Despite discarding question generation using the T5 transformer, \ours KMeans and the original CAQA performs similarly. Interestingly, we observe that CAQA\textsuperscript{*} outperforms the original CAQA in HotpotQA, suggesting that the distribution-aware sampling and iterative pseudo-processing can effectively improve the adaptation performance.

\subsection{Cluster Number in \ours KMeans Classification}
To study the influence of cluster number in unsupervised question clustering for \ours, we initialize KMeans algorithm with different number of clusters. Then we perform \ours KMeans adaptation on HotpotQA and SearchQA to examine the influence of cluster number.

\begin{table}[h]
  \centering
  \begin{tabular}{cccc}
  \hline
  \multirow{2}{*}{Cluster Number} & \textbf{HotpotQA} & \textbf{SearchQA} \\
                                   & EM / F1                 & EM / F1           \\ \hline
  3      & 49.82/65.73          & 30.34/39.49 \\
  5      & 49.58/65.78          & \textbf{32.19/41.27} \\
  7      & 49.87/65.66          & 30.51/39.25 \\ 
  9      & \textbf{50.45/66.14} & 29.59/38.14 \\ \hline
  \end{tabular}
  \caption[\ours performance w/ random sampling]{QC4QA adaptation performance for different numbers of KMeans clusters.}
  \label{tab:cluster-num}
\end{table}

Results are presented in \Cref{tab:cluster-num}, we observe performance drops when we reduce number of clusters from the default of 5. Surprisingly, the performance on HotpotQA grows consistently with increasing number of clusters. A potential explanation for such improvements is that fine-grained question classification is more helpful for complex multi-hop QA datasets.

\subsection{Human Annotation}
\begin{table}[h]
  \centering
  \begin{tabular}{lccc}
  \hline
  \multirow{2}{*}{\textbf{Training Method}} & \multicolumn{1}{c}{\textbf{Daily Mail}} & \textbf{NewsQA} \\
                                    & EM / F1       & EM / F1        \\ \hline
  0 Annotation    & 36.43/45.85     & 44.86/61.40    \\
  5k Annotations  & 48.04/56.27     & 45.71/62.21    \\
  10k Annotations & 55.37/61.95     & 47.17/63.46    \\ 
  20k Annotations & 66.83/72.18     & 48.72/64.92    \\ \hline
  \end{tabular}
  \caption[Semi-supervised adaptation performance]{Semi-supervised adaptation performance with \ours.}
  \label{tab:human-annotation}
\end{table}

We also study the influence of human annotations by introducing labeled target examples. We present the results on Daily Mail and NewsQA in \Cref{tab:human-annotation}. We observe that human annotations improve the adaptation performance in general. With the increasing amount of annotations, the performance gains of \ours rise rapidly and then stay steady. In both cases, introducing limited annotations can significantly improve model performance. The results indicate that the introduction of even limited amount of annotations helps QA systems reach comparable magnitude of supervised results.

\begin{table*}[t]
\centering
\begin{tabular}{p{0.95\linewidth}}
\hline
\multicolumn{1}{c}{TREC classification examples} \\ \hline
\textbf{Q1}: Judges in @placeholder and Oregon this week overturn marriage bans. $\rightarrow$ \textbf{DESC} \\
\textbf{Q2}: Spain international Mata close to joining English club @placeholder. $\rightarrow$ \textbf{ENTY} \\
\textbf{Q3}: The Surprise will be sold in 120 @placeholder stores, costing 1.75 for four? $\rightarrow$ \textbf{ENTY} \\ 
\textbf{Q4}: School bus drivers union will strike wednesday if it doesn't reach deal with @placeholder. $\rightarrow$ \textbf{HUM} \\
\textbf{Q5}: Serial killer Israel Keyes may have killed missing @placeholder woman. $\rightarrow$ \textbf{HUM} \\
\textbf{Q6}: Which is the latest version of corel draw? $\rightarrow$ \textbf{ENTY} \\ 
\textbf{Q7}: Who did say South Africa did not issue a visa on time? $\rightarrow$ \textbf{HUM} \\
\textbf{Q8}: Census bureaus are hiring people from where? $\rightarrow$ \textbf{LOC} \\
\textbf{Q9}: How long was the lion's longest field goal? $\rightarrow$ \textbf{NUM} \\ 
\textbf{Q10}: Musician and satirist Allie Goertz wrote a song about the "The Simpsons" character Milhouse, who Matt Groening named after who? $\rightarrow$ \textbf{HUM} \\
\textbf{Q11}: To whom did the Virgin Mary allegedly appear in 1858 in Lourdes France? $\rightarrow$ \textbf{HUM} \\ 
\textbf{Q12}: When did the Scholastic Magazine of Notre dame begin publishing?  $\rightarrow$ \textbf{NUM} \\ 
\textbf{Q13}: The Basilica of the Sacred heart at Notre Dame is beside to which structure? \textbf{ENTY} \\ 
\textbf{Q14}: How often is Notre Dame's the Juggler published? \textbf{NUM} \\ 
\textbf{Q15}: Where is the headquarters of the Congregation of the Holy Cross? \textbf{LOC} \\ 
\textbf{Q16}: What is the oldest structure at Notre Dame? \textbf{ENTY} \\ 
\textbf{Q17}: Which organization declared the First Year of Studies program at Notre Dame "outstanding"? \textbf{HUM} \\ 
\textbf{Q18}: The College of Science began to offer civil engineering courses beginning at what time at Notre Dame? \textbf{HUM} \\ 
\textbf{Q19}: In what year was the College of Engineering at Notre Dame formed? \textbf{NUM} \\ 
\textbf{Q20}: Which prize did Frederick Buechner create? \textbf{ENTY} \\ 
\textbf{Q21}: What was the amount of children murdered? \textbf{NUM} \\ 
\textbf{Q22}: Where was one employee killed? \textbf{HUM} \\ 
\textbf{Q23}: What happened in Chad? \textbf{DESC} \\
\textbf{Q24}: What did one of John II's replacements do in captivity? \textbf{ENTY} \\ 
\textbf{Q25}: Who threw the first touchdown pass of the game? \textbf{HUM} \\ 
\textbf{Q26}: Which player scored touchdowns running and receiving? \textbf{HUM} \\ 
\textbf{Q27}: What all field goals did Olindo Mare make? \textbf{ENTY} \\
\textbf{Q28}: Which team had a safety scored on them in the first half? \textbf{HUM} \\ 
\textbf{Q29}: What was the difference between the role of blacks and whites in the draft? \textbf{DESC} \\
\textbf{Q30}: What was burned last: city of Ryazan or suburbs of Moscow? \textbf{LOC} \\ 
\hline
\end{tabular}
\caption[Question classification examples]{Qualitative examples of classified questions in target datasets.}
\label{tab:qc-examples}
\end{table*}

\end{document}